\documentclass[11pt, a4paper, logo, copyright]{kunumi}
\usepackage[authoryear, sort&compress, round]{natbib}
\usepackage[]{mdframed}
\usepackage{anyfontsize}
\usepackage{listings}
\usepackage{hyperref}
\usepackage{url}
\usepackage{graphicx}
\usepackage{mathrsfs} %
\usepackage{etoolbox}
\usepackage{cleveref}
\usepackage{tcolorbox}
\usepackage{tabularx}
\usepackage{colortbl}
\usepackage{booktabs}       %
\usepackage{amsfonts}       %
\usepackage{nicefrac}       %
\usepackage{microtype}      %
\usepackage{subcaption}
\usepackage{multirow}
\usepackage{subcaption}
\usepackage{ragged2e}
\usepackage{enumitem}%
\setlist[itemize]{noitemsep, topsep=0pt}
\usepackage{enumitem,kantlipsum}
\usepackage{tabularx}
\usepackage{ragged2e} %
\usepackage{booktabs} %
\usepackage{xr} 
\usepackage{xcolor}

\newlength\savewidth

\definecolor{baselinecolor}{HTML}{d6eaf8}

\definecolor{mygray}{gray}{0.4}

\definecolor{darkgreen}{rgb}{0, 0.5, 0}

\AtBeginEnvironment{tcolorbox}{\tiny}

\usepackage{adjustbox}
\usepackage[utf8]{inputenc} 
\usepackage[T1]{fontenc}    
\usepackage{xcolor}         
\usepackage{xspace}
\usepackage[english]{babel}    
\usepackage{graphicx}          
\usepackage{amsmath}           
\usepackage{amssymb}           
\usepackage{multirow}          
\usepackage{rotating}          
\usepackage{soul}              
\usepackage{appendix}          
\usepackage[acronym]{glossaries}  
\usepackage{dsfont}
\usepackage{enumitem}
\usepackage{fancyhdr}
\usepackage{colortbl}
\usepackage{makecell}
\usepackage{threeparttable}
\usepackage{algorithm}
\usepackage{algpseudocode}
\usepackage[dvipsnames]{xcolor}

\usepackage{float}

\title{Continual Learning for Sequential Personalization of Small Language Models: A Stability Monitoring Analysis}
\correspondingauthor{Thomas S. Paula (thomas.paula@edu.pucrs.br), Lucas S. Kupssinskü (lucas.kupssinsku@pucrs.br) and Rodrigo C. Barros (rodrigo.barros@pucrs.br)}

\author[1]{Thomas S. Paula}
\author[1]{Lucas S. Kupssinskü}
\author[1,2]{Rodrigo C. Barros}

\affil[1]{MALTA, Machine Learning Theory and Applications Lab, PUCRS, Porto Alegre, Brazil}
\affil[2]{Kunumi Institute, Brazil}

\newif\ifonecolumn
\onecolumntrue %
\begin{abstract}
\vspace{-1em}
Small Language Models (SLMs) are increasingly being considered for deployment on edge devices such as laptops, enabling private, low-latency, and locally personalized applications. However, personalization requires models to adapt over time to evolving user- or task-specific data, placing them in a continual learning setting. This creates the risk of catastrophic forgetting, where learning new information degrades performance on previously learned tasks or broader model capabilities. Recent benchmarks such as TRACE have shown that continual fine-tuning can significantly degrade the general abilities of aligned large language models. In this work, we present a study for sequential LoRA personalization of SLMs. We save model checkpoints after each adaptation stage and evaluate them on current tasks, previously seen tasks, and a fixed reference set. This checkpoint-level protocol enables us to monitor task performance, forgetting, and reference set drift over time. We show that lightweight reference set distributional diagnostics can reveal model-specific instability patterns during sequential LoRA personalization of SLMs, including cases where task-level metrics alone hide harmful adaptation. We hope this can highlight new research avenues for monitoring stability of SLMs in a continual learning setting.
\end{abstract}

\begin{document}

\maketitle

\section{Introduction}

Large Language Models (LLMs) demonstrated strong performance across a variety of natural language processing tasks, becoming pervasive across applications and domains.
Although most of the state-of-the-art models run on the cloud, there is a recent effort on deploying LLMs to the edge, motivated by privacy, personalization, and dedicated compute such as Neural Processing Units (NPUs)~\citep{xu2025fast}.
In order to do so, techniques such as quantization, pruning, and distillation have been employed to compress the models to a size that can fit on edge devices.
The research and industry communities have called these models Small Language Models (SLMs), which allow significant computational savings in pre-training, fine-tuning, and inference with reduced memory and storage needs due to the smaller parameters count~\citep{10.1145/3768165}.

SLMs are increasingly relevant for edge and personal computing scenarios, where local inference can provide lower latency, improved privacy, and reduced dependence on cloud services.
In these settings, personalization becomes a natural requirement: models may need to adapt to user- or task-specific preferences, domains, writing styles, or any other data distributions over time.
However, sequential personalization introduces a continual learning problem.
A model adapted to a new task or user distribution may degrade on previously learned tasks or on broader reference behavior.
This phenomenon is commonly studied as catastrophic forgetting~\citep{french1999catastrophic}.
In deployed SLM systems, the system also must remain useful, predictable, and stable across a sequence of updates.

Recent benchmarks such as TRACE~\citep{wang2023trace} show that continual fine-tuning can degrade aligned LLMs across sequential tasks and broader capabilities, including general ability and instruction following.
TRACE evaluates models over diverse sequential tasks and uses metrics such as Overall Performance (OP) and Backward Transfer (BWT), while also introducing broader deltas for general ability, instruction following, and safety.
However, most recent papers building on top of such benchmarks do not focus on edge scenarios, where the size of the models and the amount of available memory make it even more challenging.

Motivated by this landscape, we study sequential personalization of SLMs from a stability monitoring perspective.
Our focus is to evaluate how an SLM behaves when adapted sequentially using a lightweight parameter-efficient method.
In particular, we ask whether checkpoint-level evaluation can expose useful signals about acquisition, forgetting, transfer, and stability during personalization.
We present a monitoring protocol inspired by the evaluation logic of TRACE, but focused on a smaller number of tasks and stability.
A base SLM is adapted sequentially across a small number of tasks using LoRA~\citep{hu2022lora}.
After each adaptation stage, the resulting checkpoint is evaluated not only on the current task, but also on previously seen tasks, future tasks, and a fixed reference set. This produces a checkpoint-by-task matrix that allows us to track how model behavior evolves over time.
This paper makes three contributions~\footnote{Code is available at: \url{https://github.com/tspthomas/slm_stability_cl}}:

\begin{enumerate}
    \item We operationalize sequential personalization of SLMs as a checkpoint-level stability monitoring problem. Rather than evaluating only final-task performance, we track how each adaptation stage affects current-task acquisition, retention on previous tasks, transfer to future tasks, and behavior on a fixed reference set.

    \item We introduce a lightweight monitoring protocol for cumulative LoRA personalization of SLMs. It combines a checkpoint-by-task evaluation matrix with standard continual learning metrics and reference set diagnostics.

    \item We provide an empirical analysis across three SLMs and two task orders, showing that predictive drift is jointly influenced by model identity and task identity. In particular, NumGLUE-CM consistently induces the largest KL-divergence increase, while Gemma exhibits the greatest drift sensitivity.

\end{enumerate}

Overall, the study positions reference set monitoring as a practical diagnostic tool for long-running SLM personalization. While not intended as a comprehensive benchmark or a new continual learning algorithm, results show that checkpoint-level stability signals can expose failure modes that are not visible from final accuracy alone.

\section{Background}

\subsubsection{Continual Learning and Evaluation of Language Models.}
Continual learning is a machine learning approach for learning continuously, accumulating the knowledge learned in previous tasks, and using it to help future learning~\citep{chen2018lifelong}.
It breaks the traditional assumption that the data distribution remains static, connecting better with real-world applications where data is non-stationary~\citep{YANG2026108226}.
The key objectives of CL are to avoid catastrophic forgetting and facilitate knowledge transfer~\citep{chen2026continual}.
In LLMs, this phenomena is quite important because continuous fine-tuning may degrade general capabilities learned during pre-training or later stages.
TRACE~\citep{wang2023trace} is a benchmark designed to evaluate CL in LLMs.
The authors show that continual fine-tuning can degrade aligned LLMs across sequential tasks and broader capabilities, including general ability and instruction following.
Existing CL benchmarks for LLMs focus primarily on measuring task performance and capability degradation, whereas our work focuses on checkpoint-level stability monitoring for cumulative LoRA personalization of SLMs.

\subsubsection{Parameter-Efficient Adaptation for Personalization.}

Fine-tuning all parameters of large language models is prohibitive for most use cases due to the memory and processing requirements.
Parameter Efficient Fine-Tuning (PEFT) emerged as a promising technique, where less parameters are trained, in a memory-efficient manner~\citep{houlsby2019parameter}. 
Low-Rank Adapdation (LoRA) is a well-known parameter-efficient fine tuning method, which adapts LLMs to downstream tasks without the large computational cost as a full fine-tuning approach~\citep{hu2022lora}.
Multiple LoRA variants emerged recently, including QLoRA~\citep{NEURIPS2023_1feb8787}, which combines 4-bit quantization with LoRA to enable memory efficient fine-tuning of LLMs; and O-LoRA~\citep{wang-etal-2023-orthogonal}, which extends LoRA for continual learning, having a new LoRA module for each task in a temporal sequence.
These works demonstrate efficient approaches that are suitable for on-device personalization, but they do not study stability of the adaptation process over time.

\subsubsection{Stability and Drift Monitoring During Adaptation}
General continual learning metrics emphasize task accuracy, forgetting, and transfer, which not always reflect why models are degrading.
In sequential personalization tasks, a model may improve on the current task while drifting away from its original behavior --- it may have lost capabilities learned during pre-training or alignment.
Therefore, it is paramount to understand if the model changed its overall behavior and how much it is drifting away from the previously learned tasks.
We hypothesize that a fixed reference set combined with KL divergence, entropy, and margin is a simple stability analysis tool.
Our work uses those stability signs to monitor different model checkpoints, as a way to check whether they can reveal instability during sequential adaptation and understand model degradation.

\section{Stability Monitoring Protocol}

\subsection{Problem Setting}

As a base for our experiments, we defined our sequential personalization setting following a continual learning setup with a sequence tasks:
\[
\mathcal{T} = (T_1, T_2, \ldots, T_K),
\]
where \(K\) is the number of personalization stages.
Each task \(T_k\) is associated with a training set and an evaluation set:
$D_k^{\mathrm{train}} = \{(x_n, y_n)\}_{n=1}^{N_k^{\mathrm{train}}}$, $D_k^{\mathrm{eval}} = \{(x_n, y_n)\}_{n=1}^{N_k^{\mathrm{eval}}}$.
The training set \(D_k^{\mathrm{train}}\) is used only when adapting the model at stage \(k\).
The evaluation set \(D_k^{\mathrm{eval}}\) is never used for training and is used to evaluate task-specific performance across checkpoints.

In addition to task-specific evaluation sets, we define a fixed reference set $R = \{(x_n^R, y_n^R)\}_{n=1}^{N_R}$.
The reference set \(R\) is held fixed throughout the experiment and is never used for training. While \(D_k^{\mathrm{eval}}\) measures performance on task \(T_k\), the reference set \(R\) is used to monitor whether sequential personalization impacts overall model performance over time.
In summary:
\[
D_k^{\mathrm{train}} \cap D_k^{\mathrm{eval}} = \emptyset,
\]
\[
R \cap D_k^{\mathrm{train}} = \emptyset
\quad \forall k \in \{1,\ldots,K\},
\]
\[
R \cap D_k^{\mathrm{eval}} = \emptyset
\quad \forall k \in \{1,\ldots,K\}.
\]

Different from a continual learning setting, we are not interested in optimizing model's accuracy over time.
We are rather focused on understanding the dynamics and stability of the model during such a process.

\subsection{Model Checkpoints and Evaluation}
\label{subsec:model_checkpoints}

Let \(\theta_0\) denote the initial SLM before personalization (also called the base or reference model interchangeably).
After adapting the model on task \(T_k\), we obtain checkpoint \(\theta_k\).
The full checkpoint sequence is: $\Theta = (\theta_0, \theta_1, \ldots, \theta_K).$
At stage \(k\), the checkpoint $\theta_{k-1}$ is fine-tuned only using the current task training set:
\[
\theta_k = \mathrm{train}(\theta_{k-1}, D_k^{\mathrm{train}}).
\]

Let $a_{i,j}$ denote the performance of checkpoint \(\theta_i\) in the evaluation set of task \(T_j\), namely \(D_j^{\mathrm{eval}}\). The checkpoint-by-task evaluation matrix is:
\[
A =
\begin{bmatrix}
a_{0,1} & a_{0,2} & \cdots & a_{0,K} \\
a_{1,1} & a_{1,2} & \cdots & a_{1,K} \\
\vdots & \vdots & \ddots & \vdots \\
a_{K,1} & a_{K,2} & \cdots & a_{K,K}
\end{bmatrix}.
\]

Rows correspond to model checkpoints and columns correspond to task evaluation sets. We followed the convention from~\citep{NIPS2017_f8752278}, which simplifies interpration and computing standard metrics.
The task performance score can be defined as:
\[
a_{i,j}
=
\frac{1}{|D_k^{\mathrm{eval}}|}
\sum_{(x,y) \in D_k^{\mathrm{eval}}}
s(f_{\theta_i}(x), y),
\]
where \(f_{\theta_i}(x)\) is the output generated by checkpoint \(\theta_i\), \(y\) is the target answer, and \(s(\cdot,\cdot)\) is a task-specific scoring function.
In our case, the scoring function is a simple accuracy: in case the model predicts the correct choice for a multiple choice problem or the correct numeric value for a math-related problem, that is considered a correct answer; otherwise, we consider it incorrect.

\subsection{Metrics}
\label{subsec:metrics}

\subsubsection{Continual Learning.}

Following~\citep{NIPS2017_f8752278}, we employed standard continual learning metrics that explore different aspects of the learning process.
More specifically, we used the following metrics.

\paragraph{Average Accuracy.}
Average Accuracy after learning the full task sequence is:
\[
\mathrm{ACC}
=
\frac{1}{K}
\sum_{j=1}^{K}
a_{K,j}.
\]

\paragraph{Backward Transfer.}
Backward Transfer measures the effect of later training on previously learned tasks:
\[
\mathrm{BWT}
=
\frac{1}{K-1}
\sum_{j=1}^{K-1}
\left(
a_{K,j} - a_{j,j}
\right).
\]

\paragraph{Forward Transfer.}
Forward Transfer measures the effect of earlier training on future tasks before they are learned:
\[
\mathrm{FWT}
=
\frac{1}{K-1}
\sum_{j=2}^{K}
\left(
a_{j-1,j} - a_{0,j}
\right).
\]

In addition to standard continual learning metrics, we report two diagnostic
quantities motivated by the stability-monitoring objective of this study.
First, we report the diagonal entries $a_{k,k}$, corresponding to performance
on each task after it is learned. We summarize these values as
Learning Accuracy (LA):
$\mathrm{LA} = \frac{1}{K} \sum_{k=1}^{K} a_{k,k}$.
Second, we report the immediate adaptation gain: $\Delta_{\mathrm{adapt}}(k) = a_{k,k} - a_{k-1,k}$
This quantity is not intended as a replacement for standard continual learning
metrics. Instead, it helps diagnose whether each personalization stage improves
performance on the current task before analyzing retention, transfer, and
reference-set stability.

\subsubsection{Stability.} We implemented different metrics to explore how stable the adaptation process is over time.
We start with our fixed reference set $R$.
For each reference example $x \in R$ and checkpoint $\theta_k$, we compute the
next token predictive distribution as:
\[
p_k(y \mid x)
=
\mathrm{softmax}\left(z_k(x)\right)_y,
\]
where $z_k(x)$ are the next token logits produced by checkpoint $\theta_k$.

\paragraph{KL Divergence.}
Our main metric is KL divergence~\citep{shlens2014noteskullbackleiblerdivergencelikelihood}, which we compute relative to the base model $\theta_0$.
To measure the drift from the base model, we compute
\[
D_{\mathrm{KL}}(p_k \Vert p_0)(x)
=
\sum_{y \in \mathcal{V}}
p_k(y \mid x)
\left[
\log p_k(y \mid x) - \log p_0(y \mid x)
\right],
\]
where $p_0(y \mid x)$ is the next token distribution of the base model.
The reference set $R$ KL divergence is then
\[
\overline{D}_{\mathrm{KL}}(p_k \Vert p_0)
=
\frac{1}{|R|}
\sum_{x \in R}
D_{\mathrm{KL}}(p_k \Vert p_0)(x).
\]

For the base checkpoint, the KL divergence is defined as $\overline{D}_{\mathrm{KL}}(p_0 \Vert p_0) = 0$.
For batched inference, prompts are left-padded and the next-token logits are selected at the rightmost position whose attention-mask value is one.
The base distribution is obtained from the same model with the cumulative LoRA adapter disabled.

\paragraph{Entropy.}
Entropy is reported as the mean over of the reference set $R$:
\[
\overline{H}_k
=
\frac{1}{|R|}
\sum_{x \in R} H_k(x),
\]
where $H_k(x) = -\sum_{y \in \mathcal{V}} p_k(y \mid x)\log p_k(y \mid x)$ and 
$\mathcal{V}$ is the model vocabulary.
To compare checkpoints relative to the base model, we also report $\Delta$ Entropy ($\Delta H_k = \overline{H}_k - \overline{H}_0$).
Entropy is commonly used in uncertainty estimation to assess model confidence~\citep{lakshminarayanan2017simple}.

\paragraph{Margin.}
Finally, the last stability-related metric we compute is margin.
Let $y_{(1)}$ and $y_{(2)}$ denote the most likely and second-most likely
next tokens under $p_k(\cdot \mid x)$.
We define the margin as
\[
m_k(x)
=
\log p_k(y_{(1)} \mid x)
-
\log p_k(y_{(2)} \mid x).
\]
The reference set margin is
\[
\overline{m}_k
=
\frac{1}{|R|}
\sum_{x \in R} m_k(x),
\]
and the change relative to the base checkpoint is $\Delta m_k = \overline{m}_k - \overline{m}_0$.
This definition of margin is simple and useful, but it is not new.
Previous work on the active learning literature and recent work have been exploring its use in LLMs~\citep{ramirez2024optimising}, and we adapt its use for understanding stability.

\section{Experimental Setup}






\subsubsection{Datasets}

In order to explore both the continual learning and the stability scenarios, we leverage the TRACE benchmark~\citep{wang2023trace}.
TRACE is comprised of eight main datasets, which cover different tasks.
Since our focus is on constrained scenarios, we selected three representative datasets for our experiments: FOMC~\citep{shah-etal-2023-trillion}, ScienceQA~\citep{NEURIPS2022_11332b6b}, and NumGLUE~\citep{mishra-etal-2022-numglue}.

FOMC is a hawkish-dovish classification task, which aims to classify central bank communications as hawkish, dovish, or neutral.
Given the market condition, we need to classify central bank communications (meeting minutes, speeches) as hawkish (favoring high-interest rates to fight inflation), dovish (favoring lower rates to stimulate growth), or neutral.
ScienceQA is a question and answer dataset with questions from elementary and high school science curricula, including natural science, social science, and language science examples.
Finally, NumGLUE is a mathematical reasoning dataset, focused on arithmetic reasoning abilities, where the model must provide the answer of arithmetic problems.

All the datasets we used are from TRACE's original code repository~\footnote{Available at \url{https://github.com/BeyonderXX/TRACE}}.
Table~\ref{tab:trace_datasets} below summarizes the datasets used (part of the information is taken from~\citep{liao2025data}).
We opted for the smaller sample of training data to constrain the scenarios for a more realistic edge training setting.

\begin{table}[ht!]
\centering
\caption{Datasets used in the TRACE-style continual learning evaluation.}
\label{tab:trace_datasets}
\begin{tabular}{l l c c c c}
\toprule
Dataset & Domain & Avg. Length & Metric & Language & Train Examples \\
\midrule
ScienceQA  & Science     & $210$ & Accuracy & English & $500$ \\
FOMC       & Finance     & $51$  & Accuracy & English & $500$ \\
NumGLUE-CM & Mathematics & $32$  & Accuracy & English & $500$ \\
\bottomrule
\end{tabular}
\end{table}

\subsubsection{Models}

We selected three recent state-of-the-art SLMs, that represent different families of models.
A key criterion was that all of them should have $\leq1$ billion parameters.
Therefore, we selected Qwen 3.5 0.8B~\citep{qwen3.5}, Llama 3.2 1B Instruct~\citep{llama32_1b_instruct_modelcard,dubey2024llama3}, and Gemma 3 1B IT~\citep{gemmateam2025gemma3technicalreport}.
Table~\ref{tab:models} summarizes them.
For now onwards, we will refer to them as simply Qwen, Llama, and Gemma.

\begin{table}[ht!]
\centering
\caption{Summary of the models used in our experiments.}
\label{tab:models}
\begin{tabular}{lll}
\toprule
Model & \makecell{Number of\\Parameters} & \makecell{Source\\(Hugging Face)} \\
\midrule
Qwen 3.5 0.8B & $0.8$B & \texttt{Qwen/Qwen3.5-0.8B} \\
Llama 3.2 1B Instruct & $1.0$B & \makecell[l]{\texttt{meta-llama/Llama-3.2-1B-Instruct}} \\
Gemma 3 1B IT & $1.0$B & \texttt{google/gemma-3-1b-it} \\
\bottomrule
\end{tabular}
\end{table}

\subsubsection{Implementation}
We leveraged Transformers~\citep{wolf-etal-2020-transformers}, PEFT~\citep{peft}, and PyTorch~\citep{paszke2019pytorchimperativestylehighperformance} libraries for implementing all our code.
All experiments were conducted on Titan X with $11$GB, with CUDA $11.8$.
All the results are reported as an average of $3$ runs with different random seeds ($33$, $42$, and $123$).
The setup for all the models is exactly the same, including any task-specific prompts.
We plan to make our code available.

\subsubsection{Training Procedure}
We adopt a sequential personalization setup with three tasks ordered as $T_1$, $T_2$, and $T_3$, which are FOMC, ScienceQA, and NumGLUE-CM, respectively.
The base instruction-tuned SLMs are first evaluated before any fine-tuning step (stage), resulting in the checkpoint $\theta_0$.
We apply LoRA to the base model and train a single cumulative adapter across the task sequences, instead of training a separate adapter for each task.
We use $r=8$, $alpha=16$, a dropout of $0.05$, and no bias for all experiments.
The target for LoRA is ``all-linear'' for all models.

All models are trained for $1$ epoch, with a \emph{batch size} $=2$, \emph{gradient accumulation steps} $=8$, \emph{learning rate} $=5e^{-5}$, and a \emph{max length} $=512$.
We adopted the AdamW~\citep{loshchilov2017decoupled} optimizer.
The optimizer is reset at the beginning of each task to avoid carrying optimizer state across tasks, while the model weights continue from the previous checkpoint.
We did not apply any hyperparameter search or exploratory exercises for determining the best parameters since our objective is not to maximize accuracy.
Although these values might be considered conservative, we wanted to have a scenario that limits the amount of updates to carefully study the stability phenomenon.

The training process happens in a sequential fashion: after training on task $T_k$, the resulting model checkpoint is denoted as $\theta_k$.
At each stage, only the current task's training set is used for fine-tuning.
We do not employ any replay techniques.
The same prompt format and answer normalization rules are used across training and evaluation to avoid template mismatch.
For multiple-choice tasks, training targets are normalized to the option letter only. For numerical tasks, targets are normalized to the final numeric answer.

\subsubsection{Evaluation Protocol}

After each checkpoint $\theta_k$ (which includes the base checkpoint $\theta_0$), we evaluate the model on all the defined tasks $T_1$, $T_2$, and $T_3$.
This produces the checkpoint-by-task accuracy matrix, explained in Section~\ref{subsec:model_checkpoints}, where each entry $a_{k,i}$ is the accuracy of checkpoint $\theta_k$ in task $T_i$.
At each checkpoint, we perform:

\begin{itemize}
    \item Evaluation of all tasks: Compute current checkpoint's performance on all tasks after training in that step. Basically, $\theta_2$ is the checkpoint after we  trained on task $T_2$. $\theta_2$ is evaluated in the test set of each task ($T_1$, $T_2$, $T_3$).
    \item Evaluation of the reference set: Compute current checkpoint's performance on our shared reference set in that step.
    \item Evaluation of the stability: Compute stability metrics based on the reference set, comparing current checkpoint's performance $\theta_k$ to the base model $\theta_0$.
\end{itemize}

In all cases, task-level performance is evaluated with exact-match accuracy after we performed answer normalization -- a collection of regular expressions that extract only the final answer from the model's output.
Multiple-choice predictions are normalized to option labels, and numerical predictions are normalized by extracting the final numeric answer.
The reference set is kept fixed across all checkpoints and seeds, and is not used for training.
For all evaluation runs, we use a \emph{batch size} of $4$ and set \emph{max tokens} to $256$, \emph{temperature} to $0.0$, and \emph{top\_p} to $1.0$.
Final task performance is reported on the test sets, while the fixed reference set is used only for stability monitoring.

In the case of continual learning metrics, we adopted the following metrics: Overall Performance (OP) is computed as the average accuracy over tasks observed so far; Backward Transfer (BWT) measures whether performance on previous tasks changes after learning later tasks; Forward Transfer (FWT) measures whether training on earlier tasks changes performance on future tasks before those tasks are learned;
Learning Accuracy (LA) summarizes the diagonal entries $a_{k,k}$, corresponding to the performance on each task immediately after it is learned; and immediate adaptation gain $\Delta_{\mathrm{adapt}}(k)$ measures the improvement on task $T_k$ from the previous checkpoint to the checkpoint after training on $T_k$.

For stability, we implemented simple, known metrics, aiming to measure the distributional shift from the base model and understand whether the adapted models are drifting.
We adopted KL divergence to measure how much a given checkpoint $\theta_k$'s next token distribution has drifted from the base checkpoint $\theta_0$.
Entropy, which is commonly used as an uncertainty signal, is calculated directly from checkpoint $\theta_k$'s next token probabilities.
We then report $\Delta$ Entropy as the difference between $H_k$ (entropy of checkpoint $\theta_k$) and $H_0$ (entropy of checkpoint $\theta_0$.
Finally, we use margin as a way to compare how decisive the model is between its top two choices.
It measures how separated the checkpoint $\theta_k$'s most likely next token prediction is from its second most likely prediction.  Those metrics can be seen as lightweight stability diagnostics.

\section{Main Results}

\subsubsection{Task Acquisition and Retention}

Table~\ref{tab:adaptation_summary} shows a summary of the sequential adaptation.
$\Delta_{adapt}$ shows that all three models achieve largest immediate gains after being trained on FOMC, suggesting this initial step is useful across different model architectures.
NumGLUE-CM is the most harmful task, where $\Delta_{adapt}$ is negative for all model architectures.
Qwen is the most stable model, maintaining the highest final average accuracy ($0.591 \pm, 0.012$), remaining better than the base model $M_0$, even after being adapted in NumGLUE-CM.
Gemma shows the highest instability.
Despite its initial strong gain with FOMC ($+0.327$), its performance goes down after adapting on NumGLUE-CM.
For both Gemma and Llama, the final $\Delta$ vs. Base scores are negative, showing that the personalization process actually degraded their performance on those tasks when we compared to the original base model.

\begin{table}[ht!]
\centering
\caption{Sequential adaptation summary: task acquisition and immediate diagnostics (order $T_1$, $T_2$, $T_3$). Results reflect mean $\pm$ standard deviation across $3$ random seeds.}
\label{tab:adaptation_summary}
\setlength{\tabcolsep}{3pt}
\resizebox{\linewidth}{!}{
\begin{tabular}{lclcccc}
\toprule
\makecell{Model} & \makecell{Step} & \makecell{Learned\\Task} & \makecell{After Learning\\$a_{k,k}$} & \makecell{Immediate Gain\\$\Delta_{adapt}$} & \makecell{$\Delta$ vs.\\Base} & \makecell{Average\\Accuracy} \\ \midrule
Gemma & 1 & FOMC & $0.547 \pm 0.012$ & $+0.327$ & $+0.327 \pm 0.012$ & $0.522 \pm 0.011$ \\
      & 2 & ScienceQA & $0.663 \pm 0.061$ & $+0.006$ & $+0.043 \pm 0.061$ & $0.438 \pm 0.041$ \\
      & 3 & NumGLUE-CM & $0.136 \pm 0.021$ & $-0.206$ & $-0.235 \pm 0.021$ & $0.320 \pm 0.029$ \\ \midrule
Qwen  & 1 & FOMC & $0.533 \pm 0.076$ & $+0.283$ & $+0.283 \pm 0.076$ & $0.524 \pm 0.029$ \\
      & 2 & ScienceQA & $0.800 \pm 0.000$ & $+0.140$ & $+0.280 \pm 0.000$ & $0.610 \pm 0.012$ \\
      & 3 & NumGLUE-CM & $0.379 \pm 0.056$ & $-0.111$ & $+0.008 \pm 0.056$ & $0.591 \pm 0.012$ \\ \midrule
Llama & 1 & FOMC & $0.530 \pm 0.000$ & $+0.280$ & $+0.280 \pm 0.000$ & $0.537 \pm 0.018$ \\
      & 2 & ScienceQA & $0.787 \pm 0.012$ & $+0.144$ & $+0.267 \pm 0.012$ & $0.578 \pm 0.004$ \\
      & 3 & NumGLUE-CM & $0.243 \pm 0.040$ & $-0.197$ & $-0.263 \pm 0.040$ & $0.485 \pm 0.016$ \\ \bottomrule
\end{tabular}
}
\end{table}

On Table~\ref{tab:cl_metrics}, we show a summary of continual learning metrics.
Qwen is the only model that presents positive BWT, which shows that learning new tasks improved the performance on older ones.
In contrast, Gemma shows severe BWT after both ScienceQA and NumGLUE-CM, showing that the model is ``loosing'' its previous knowledge when learning new tasks.
Llama shows a stable behavior when compared to Gemma, with less negative BWT.
OP results clearly show the performance drop for both Gemma and Llama after NumGLUE-CM.
When we inspect Gemma's OP trend ($0.547 \xrightarrow[]{} 0.487 \xrightarrow[]{} 0.320$, it highlights the importance of stability monitoring, as sequential personalization can eventually degrade model performance when compared to the base model.
Finally, Qwen shows the highest FWT, especially after the second step ($0.119 \pm 0.026$).
This result suggests that Qwen might be becoming ``more generalizable'', helping it perform better on future, unlearned tasks.

\begin{table}[ht!]
\centering
\caption{Continual learning metrics across sequential checkpoints (order $T_1$, $T_2$, $T_3$).}
\label{tab:cl_metrics}
\setlength{\tabcolsep}{3pt}
\resizebox{\linewidth}{!}{
\begin{tabular}{llccc}
\toprule
\makecell{Model} & \makecell{Checkpoint} & \makecell{OP} & \makecell{BWT} & \makecell{FWT} \\ \midrule
Gemma & After FOMC & $0.547 \pm 0.012$ & --- & $0.014 \pm 0.014$ \\
      & After ScienceQA & $0.487 \pm 0.055$ & $-0.237 \pm 0.050$ & $-0.029 \pm 0.019$ \\
      & After NumGLUE-CM & $0.320 \pm 0.029$ & $-0.193 \pm 0.010$ & --- \\ \midrule
Qwen  & After FOMC & $0.533 \pm 0.076$ & --- & $0.074 \pm 0.012$ \\
      & After ScienceQA & $0.670 \pm 0.026$ & $0.007 \pm 0.075$ & $0.119 \pm 0.026$ \\
      & After NumGLUE-CM & $0.591 \pm 0.012$ & $0.030 \pm 0.023$ & --- \\ \midrule
Llama & After FOMC & $0.530 \pm 0.000$ & --- & $0.027 \pm 0.028$ \\
      & After ScienceQA & $0.647 \pm 0.008$ & $-0.023 \pm 0.021$ & $-0.066 \pm 0.019$ \\
      & After NumGLUE-CM & $0.485 \pm 0.016$ & $-0.052 \pm 0.012$ & --- \\ \bottomrule
\end{tabular}
}
\end{table}

\subsubsection{Stability Monitoring Diagnostics}

In Table~\ref{tab:stability_metrics}, we summarize the stability metrics computed relative to the base checkpoint.
We can clearly see large KL to base values for Gemma, which reach a peak of $8.022 \pm 0.243$, showing a correlation between internal distributional drift and the performance degradation observed in earlier sections.
$\Delta$ Margin is also substantially negative ($-7.795$), indicating a potential collapse in the separation between the top two predicted tokens.
In addition, Gemma's $\Delta$ Entropy increases to $+1.541$ while Qwen and Llama show mostly negative entropy changes.
Gemma’s KL to base at the first ($3.019$) is already high, while Qwen and Llama remain closer to their base distributions during the first two stages, but both exhibit large KL increases after NumGLUE-CM.
This pattern suggests that drift depends not only on the model but also strongly on the task being learned.

\begin{table}[ht!]
\centering
\caption{Reference set stability metrics computed relative to the base checkpoint.  $\Delta$ Entropy and $\Delta$ Margin are measured relative to each model's base checkpoint. 
KL to Base measures distributional drift from the base model on the fixed reference set. }
\label{tab:stability_metrics}
\begin{tabular}{llccc}
\toprule
\makecell{Model} & \makecell{Checkpoint} & \makecell{$\Delta$ Entropy} & \makecell{$\Delta$ Margin} & \makecell{KL to Base} \\ \midrule
Gemma & After FOMC & $+0.565$ & $-6.483$ & $3.019 \pm 0.051$ \\
      & After ScienceQA & $+0.838$ & $-7.672$ & $3.683 \pm 0.237$ \\
      & After NumGLUE-CM & $+1.541$ & $-7.795$ & $8.022 \pm 0.243$ \\ \midrule
Qwen  & After FOMC & $-0.304$ & $-0.168$ & $0.469 \pm 0.232$ \\
      & After ScienceQA & $-0.414$ & $-0.061$ & $0.371 \pm 0.045$ \\
      & After NumGLUE-CM & $-0.158$ & $-0.538$ & $3.269 \pm 0.071$ \\ \midrule
Llama & After FOMC & $-0.696$ & $-0.411$ & $0.489 \pm 0.042$ \\
      & After ScienceQA & $-0.909$ & $+0.257$ & $0.663 \pm 0.026$ \\
      & After NumGLUE-CM & $-0.433$ & $+0.142$ & $3.801 \pm 0.122$ \\ \bottomrule
\end{tabular}
\end{table}

\subsubsection{Association Between Predictive Drift and Task Performance}

\begin{figure}[ht]
    \centering
    \includegraphics[width=0.95\textwidth]{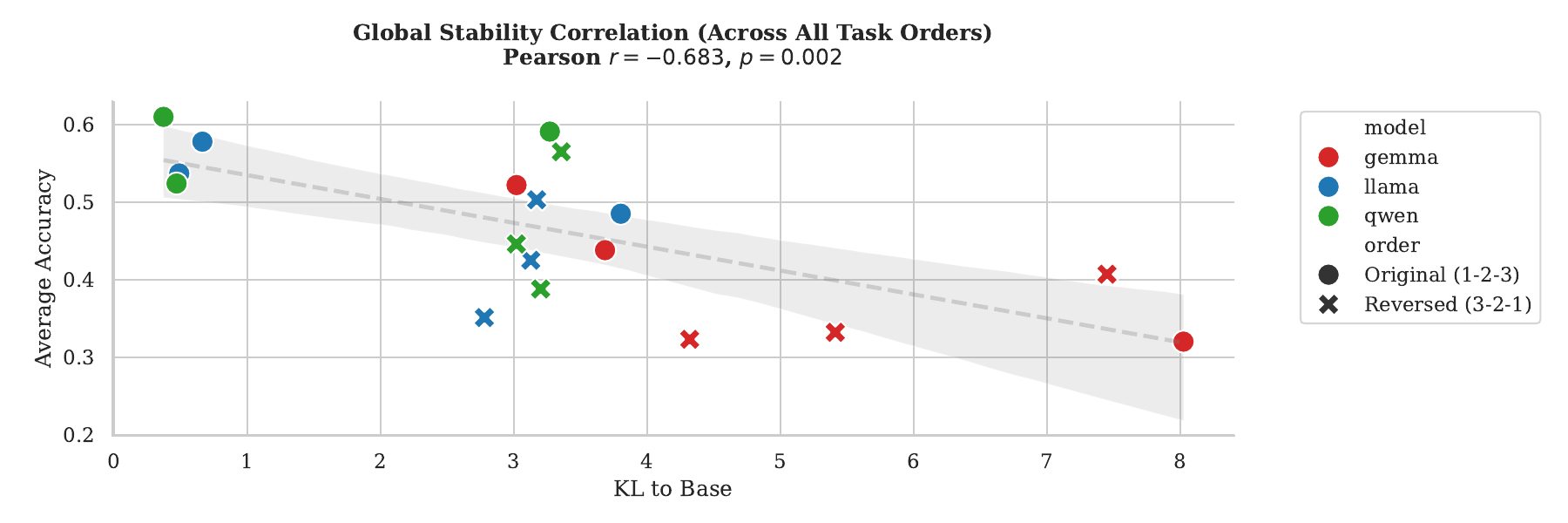}
    \caption{Global correlation between predictive-distribution drift ($D_{KL}$) and average accuracy across all models and task sequences.}
    \label{fig:global_correlation}
\end{figure}

In Figure~\ref{fig:global_correlation}, we plot the average accuracy of each step of each model considering two cases: original task order original ($T_1$, $T_2$, $T_3$) and reversed ($T_3$, $T_2$, $T_1$).
We executed the same experiments but running NumGLUE-CM first and FOMC last, with the intention of understanding the impact of placing the most challenging task to the first position. Across the 18 adapted checkpoints from both task orders, KL divergence is negatively associated with average task accuracy (\(r=-0.683,\ p= 0.002\)). This pooled result indicates that checkpoints farther from their respective base-model distributions tend to have lower average accuracy. However, the association varies substantially by order: it is strong for the original order (\(r=-0.847,\ p= 0.004\)) but weak for the reversed order (\(r=-0.286,\ p=0.455\)). We therefore interpret KL as a useful descriptive monitoring signal rather than a universal predictor of performance.


\subsubsection{Task- and Model-Dependent Drift Patterns}
\begin{figure}[ht!]
    \centering
    \includegraphics[width=\textwidth]{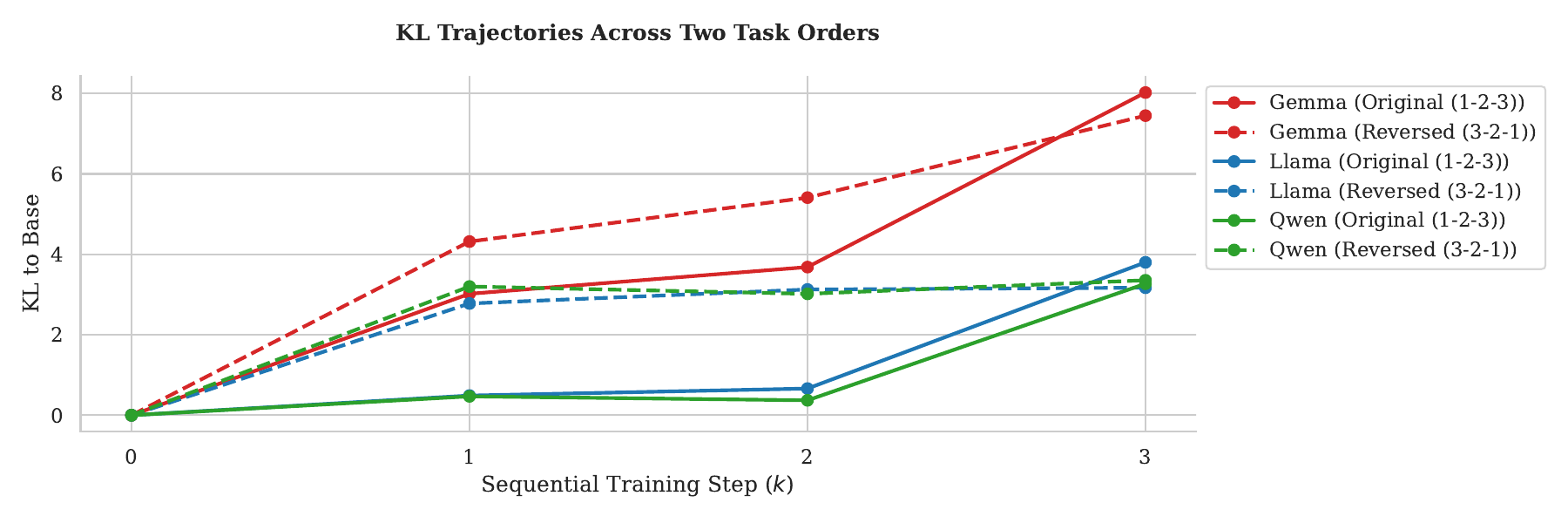}
    \caption{Internal stability signatures across varying task orders. Solid lines represent the original sequence ($T_1$, $T_2$, $T_3$) and dashed lines represent the reversed sequence ($T_3$, $T_2$, $T_1$). Early trajectories differ substantially across orders, while Gemma retains the largest final predictive-distribution drift in both orders.}
    \label{fig:order_invariance}
\end{figure}

We further expand our analysis to evaluate whether our stability findings still remain solid when we change the task order.
Figure~\ref{fig:order_invariance} shows the behavior of all models for the original and reversed task sequence.
KL trajectories differ substantially between the original and reversed task orders, particularly during the earlier adaptation stages. This rules out a strict order-invariance interpretation. Nevertheless, the final checkpoints retain a consistent broad separation: Gemma exhibits substantially greater drift than Qwen and Llama in both orders. The trajectories further show that NumGLUE-CM causes the largest KL increase for every evaluated model, whether it is learned first or last. These results suggest that drift is determined jointly by model susceptibility and the task being learned.


To validate the reliability of our monitoring protocol, Table~\ref{tab:consistency_analysis} summarizes the final results for both the original ($T_1$, $T_2$, $T_3$) and reversed ($T_3$, $T_2$, $T_1$) task sequences.
While final accuracy can vary based on the specific task order, we can see that KL to base stays relative stable across both runs, showcasing its value as a potential model stability indicator.
Final KL retains broad model-level ordering across the two sequences, although its exact magnitude remains order-sensitive.

\begin{table}[ht!]
\centering
\caption{Consistency Analysis: Final average accuracy and internal drift (KL to Base) across original ($T_1$, $T_2$, $T_3$) and reversed ($T_3$, $T_2$, $T_1$) sequences.}
\label{tab:consistency_analysis}
\setlength{\tabcolsep}{3pt}
\begin{tabular}{lcccc}
\toprule
 & \multicolumn{2}{c}{\makecell{Final Average\\Accuracy}} & \multicolumn{2}{c}{\makecell{Final KL to Base\\(Drift)}} \\ \cmidrule(lr){2-3} \cmidrule(lr){4-5}
\makecell{Model} & \makecell{Original} & \makecell{Reversed} & \makecell{Original} & \makecell{Reversed} \\ \midrule
Gemma & $0.320 \pm 0.029$ & $0.407 \pm 0.029$ & $8.022 \pm 0.243$ & $7.448 \pm 0.080$ \\
Qwen  & $0.591 \pm 0.012$ & $0.565 \pm 0.020$ & $3.269 \pm 0.071$ & $3.355 \pm 0.081$ \\
Llama & $0.485 \pm 0.016$ & $0.503 \pm 0.012$ & $3.801 \pm 0.122$ & $3.170 \pm 0.066$ \\ \bottomrule
\end{tabular}
\end{table}

\section{Conclusion}

This paper investigated the sequential personalization of Small Language Models (SLMs) using LoRA, with a particular focus on understanding model stability during the adaptation process.
We proposed a lightweight monitoring protocol that integrates a checkpoint-by-task evaluation matrix, standard continual learning metrics, and reference set diagnostics. 
Our results show that lightweight reference set diagnostics can characterize predictive-distribution changes during sequential LoRA adaptation.
KL-divergence measurements reveal that drift is jointly task- and model-dependent: NumGLUE-CM consistently induces the largest distributional shift and negative immediate adaptation gain, while Gemma exhibits the greatest sensitivity across both evaluated task orders.
Across all $18$ model-checkpoint observations from both task orders, higher KL divergence is associated with lower average accuracy, although the strength of this association varies substantially by task order. These findings support KL divergence as a useful contemporaneous diagnostic of predictive-distribution drift, but further validation is required before it can be treated as a calibrated predictor or used to define intervention thresholds.

\subsubsection{Limitations}
Despite our contributions, our work is subject to a few limitations.
First, the scale of our study was constrained by the number of evaluated tasks; a more diverse range of domains and longer task sequences would be required to draw more broadly generalized conclusions regarding stability.
Second, we focused on a specific implementation of parameter-efficient adaptation, utilizing fixed LoRA hyperparameters without exploring alternative incremental strategies or rank adjustments.
Finally, while Gemma exhibited the largest drift among the three evaluated models, further investigation is needed to pinpoint the specific internal mechanics or weight distribution shifts that drive this observed instability compared to the more robust Qwen and Llama families.

\subsubsection{Future Work}
As future work, we plan to incorporate more challenging task sequences and evaluate additional learning approaches.
We also intend to develop automated intervention strategies -- such as early-stopping or checkpoint-reverting -- triggered by KL Divergence thresholds to move toward ``self-aware'' personalization systems.
Finally, we plan to expand this study to quantized models, as they are a primary requirement for real-world edge deployments.

\subsubsection*{Acknowledgments}
This study was financed in part by the Coordination for the Improvement of Higher Education Personnel (CAPES) -- Finance Code 001; by Conselho Nacional de Desenvolvimento Científico e Tecnológico (CNPq) -- Grant Number: 443072/2024-8; and by Fundação de Amparo à Pesquisa do Estado do Rio Grande do Sul (FAPERGS) -- Grant Numbers: 25/2551-0000891-3 and 25/2551-0002824-8.
During the preparation of this work, the authors used ChatGPT and Gemini in order to proofread the text.
The authors take full responsibility for the content of this publication.

\clearpage 

{
\small
\bibliography{references.bib}
}

\end{document}